\documentclass[epj]{webofc}
\usepackage[varg]{txfonts}   
\wocname{EPJ Web of Conferences}
\woctitle{CONF12}
\begin{document}
\selectlanguage{english}
\title{The Inverse Bagging Algorithm:\\ Anomaly Detection by Inverse Bootstrap Aggregating}
%
%

\author{Pietro Vischia\inst{1}\fnsep\thanks{\email{pietro.vischia@cern.ch}} \and
        Tommaso Dorigo\inst{2}\fnsep\thanks{\email{tommaso.dorigo@pd.infn.it}}
}

\institute{Universidad de Oviedo, Spain
  \and
  INFN-Padova, Italy
}

\abstract{%
  For data sets populated by a very well modeled process and by another process of unknown probability density function (PDF), a desired feature when manipulating the fraction of the unknown process (either for enhancing it or suppressing it) consists in avoiding to modify the kinematic distributions of the well modeled one. A bootstrap technique is used to identify sub-samples rich in the well modeled process, and classify each event according to the frequency of it being part of such sub-samples. Comparisons with general MVA algorithms will be shown, as well as a study of the asymptotic properties of the method, making use of a public domain data set that models a typical search for new physics as performed at hadronic colliders such as the Large Hadron Collider (LHC).
}
\maketitle
\section{Introduction}
\label{intro}

The most popular classification algorithms based on supervised learning require a well modeled signal and a well modeled background. For the classifier to learn how to separate the two classes, it is crucial that both models are known. The case in which either signal or background has an unknown PDF is, however, acquiring importance in many classification problems that arise in the realm of particle physics, due to the fact that every passing day more and more known models are ruled out by the data. Two scenarios are mainly interesting for particle physics: a very well known background modeled from simulation, in presence of an unknown rare signal; a very well known signal modeled from simulation, contaminated by a background of origin unclear and/or not simulable. In both scenarios, it is desirable to manipulate the fraction of the unknown process, without modifying the kinematic distributions of the very well known one.

The treatment of observed data $\mathbf{X} = (X_{i}, ... X_{n})$ sampled from a probability density function $\mathit{F}$, generally proceeds by studying a statistic of the data, $R(\mathbf{X},\mathit{F})$~(e.g. the mean). It is often useful to extract the sampling distribution, by drawing many samples $\mathbf{X}_{i}$~from the population, computing the statistic $R(\mathbf{X}_{i},\mathit{F})$~for each sample $i$, and finally studying the sampling distribution of the test statistic (e.g. the distribution of the mean). Sometimes, however, it is not possible to draw additional samples $\mathbf{X}_{i}$ from the population; this could be because the PDF $\mathit{F}$~underlying the data is unknown, or the population might not be accessible (e.g. it is not possible to draw more than one sample from the stock market data for a given year), or drawing additional samples might be unfeasible or expensive (e.g. because of limited access to telescope time). In all these cases, only a single set of sampled data $\mathbf{X}$ is available.

If the population is not accessible, however, then it is possible to sample from an estimate of it, by following the plug-in principle~\cite{EfronTibshirani}; the sampled data $\mathbf{X}$~are considered as an estimate of the population, $\mathit{\hat{F}}$. Many samples $\mathbf{X^*}_{i}$~are drawn with replacement from $\mathbf{X}$, the test statistic $R^*(\mathbf{X^*}_{i},\mathit{\hat{F}})$~is computed for each $i$, and the distribution of the test statistic is studied. This procedure is referred to as {\it bootstrap}~\cite{EfronTibshirani}.
Performing the sampling with replacement (i.e. allowing for any single event to be picked for multiple samples) zeroes the covariance between the samples, that are therefore independent~\cite{Rice}

%

When data are described by using a multiple set of features, each data point can be represented as a vector, and the set of data can be written $\mathbf{\vec{X}} = (\vec{X}_{i}, ... \vec{X}_{n})$. A multivariate classifier is a function of vectorial data, i.e. a statistic of those data, in which the exact form of the function $R(\mathbf{X},\mathit{F})$ is determined by applying some optimization criteria to the test statistic computed on a training sample; the optimization criteria are chosen such that they maximize the separation between the score values for one class (signal) and the score values for the other class (background). Since such a classifier is a statistic, it is possible to estimate it by using the bootstrap method. This is performed by applying a given classification technique to many training sets obtained by bootstrap, and results in a set of independent classifiers: the final classification of each event comes from a majority vote between the individual classifications. This is a general procedure, which is applicable to nearly every classification technique. The main benefit is that the classification is less dependent on statistical fluctuations in the training sample, and it has been shown that the classification performance can improve significantly, outperforming even boosting techniques~\cite{Narsky:2005hn}.


\section{Inverse Bagging}
\label{sec:inverse_bagging}
In this paper, the test scenario considered involves a small, unknown signal on top of a very well known background. A test set of $N_{test}$ events is selected, and its composition is parametrized by a background fraction $\mathcal{B}$ close to unity. As a consequence, the total number of events populating the test sample will be composed by $\mathcal{B}N_{test}$ events from the well modeled class (background) and by $(1-\mathcal{B})N_{test}$ events from the class with properties assumed unknown (signal). A separate set of $N_{train}$ events, composed exclusively by background events, is selected as reference set for the modeling of the background features, and labeled as {\it training sample}.

From the test sample, a subset of $M<<N_{test}$ events is randomly picked. On average, the subset is expected to be populated by $\mathcal{B}\times M$ background events, but statistical fluctuations amount to a small but finite chance that all $M$ events are picked from the background. A comparison between the features of the subset of $M$ events and the background PDF estimated using the training sample can be operated, and a statistical test be devised to answer to the question ``how likely is that the M-events set is composed entirely of background events''.

Using a bootstrap procedure, a very large number of subsets of $M$ events is sampled with replacement from the test sample. Given the assumed unbalance ($\mathcal{B}$ close to unity) in the composition of the test sample, it is possible to draw as many subsets classified as background-rich as desired. For each test event $i$, two quantities are computed: the number of times $tried_{i}$~it is picked up to be part of a subset; and the number of times $ok_{i}$ the subset it is in is classified as background-like. Events are ordered according to the value of the ratio
\begin{equation}
  \label{eq:oktriedratio}
  R_{o/t} := \frac{ok_{i}}{tried_{i}}
\end{equation}

The performance of the algorithm in terms of efficiency vs. purity is then determined by removing progressively the events characterized by the largest value of the ratio.

A more sophisticated ordering principle replaces $R_{o/t}$ with the average value of the test statistic over the subsets the event belongs to. This has the advantage of not limiting the ordering to a binary classification of the subset.

The classification of the $M$-events subsets is performed by comparing each one of them to the pure-background training sample. Since $M$ is typically small, it is not advisable to rely on asymptotic approximations such as the $\chi^{2}$ method. A basic choice is the Kolmogorov-Smirnov (KS) test statistic. However, the KS test statistic is known to be insensitive to differences in the tails of the distributions, and many features used in HEP exhibit differences between signal and background mainly in the tails. The Anderson-Darling (AD) test statistic, designed to be sensitive to differences in the tails of the distributions, is a better choice. However, the KS and AD test statistics are single-dimensional tests. Zech's Energy Test (ET)~\cite{Aslan:2002cn}, based on treating the weighted distances $R$ between the multi-dimensional points as the potential energy of a set of charges, is used in order to better combine the information coming from the different feature: the weights are weighted using a weighted function that can have different forms, depending on the desired form of the pontential energy. Finally, an attempt to a multi-dimensional goodness-of-fit test is performed: this is a variation of Zech's approach, in that the weighted Euclidean distance between the multidimensional points is substituted with the nearest-neighbour distances ratio $R$. This results in enhanced power for testing localized differences between the distributions, at the price of being computationally intensive.

The performance of the algorithm is tested with respect to current methods available in the literature. A meaningful comparison is performed with respect to classification methods that do not rely on the PDF of the sample with unknown properties. Two classifiers are chosen: a Relative Likelihood approach, where the discrimination is derived from the ratio between the PDFs of the test and of the training sample; and a k-Nearest-Neighbour~\cite{Altman} approach, where the discrimination is derived from the ratio between the integrated distance between test event and training (background) events, and the integrated distance between events both belonging to the test set. Both reference methods make use of event based variables, whereas the Inverse Bagging algorithm uses subset properties to infer the classification of single events.

Such comparisons, and the validity of the Inverse Bagging algorithm, are based upon the possibility that sample-based statistics may contain more information than the set of event-based ones, which is taken here as a conjecture. Indeed, a limit might exist to the amount of information that can be extracted using a sample-based statistic, and it might be related to the amount of information that can be extracted using an event-based statistic. More studies are needed to better understand these limits.

\section{Performance tests}  

The performance of the Inverse Bagging algorithm is tested and compared with two reference methods, by using the {\sc HEPMASS} dataset~\cite{Baldi2016}, available at \texttt{http://archive.ics.uci.edu/ml/datasets/HEPMASS}. This dataset consists in: a background of simulated top quarks pair ($t\bar{t}$) production events, with top quarks decaying to a lepton-neutrino pair and a b-quark jet; and a simulated signal from an hypothetical new particle $X$ with mass $M_{X}=1000~GeV$, with exclusive decay of $X$ into a top quarks pair. This is the typical case study of a search for new physics in proton-proton collisions at the Large Hadron Collider (LHC). The kinematics of a $t\bar{t}$~pair produced in proton-proton collisions is known to a fine detail, and would be modified by the presence of an intermediate resonance. The full dataset provides low-level variables (lepton and jets four-momenta, b-tagging discriminators, etc.) and high-level variables ($M_{\ell\nu}$, the invariant mass of the lepton-neutrino pair, $M_{WWbb}$, the invariant mass of the decay products of the top quar, $W$ bosons and $b$ quarks, etc.): eight low-level variables are picked, not including the b-tagging discriminator.

To match the use case of the Inverse Bagging algorithm, in this study the signal is assumed to be unknown, and is used to populate exclusively the test sample.

\begin{figure}[h]
  \centering
  \includegraphics[width=0.45\linewidth,clip]{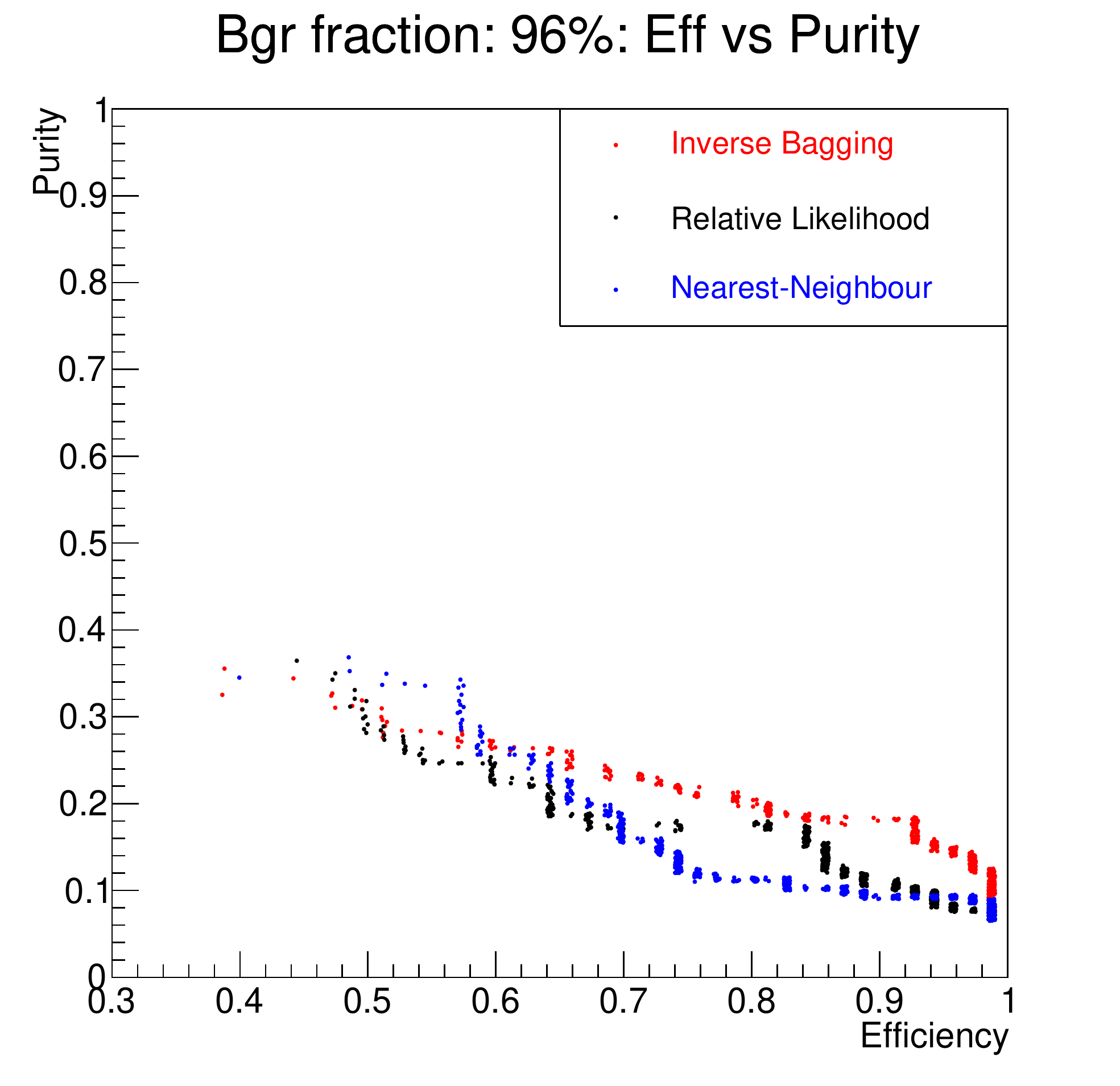}
  \includegraphics[width=0.45\linewidth,clip]{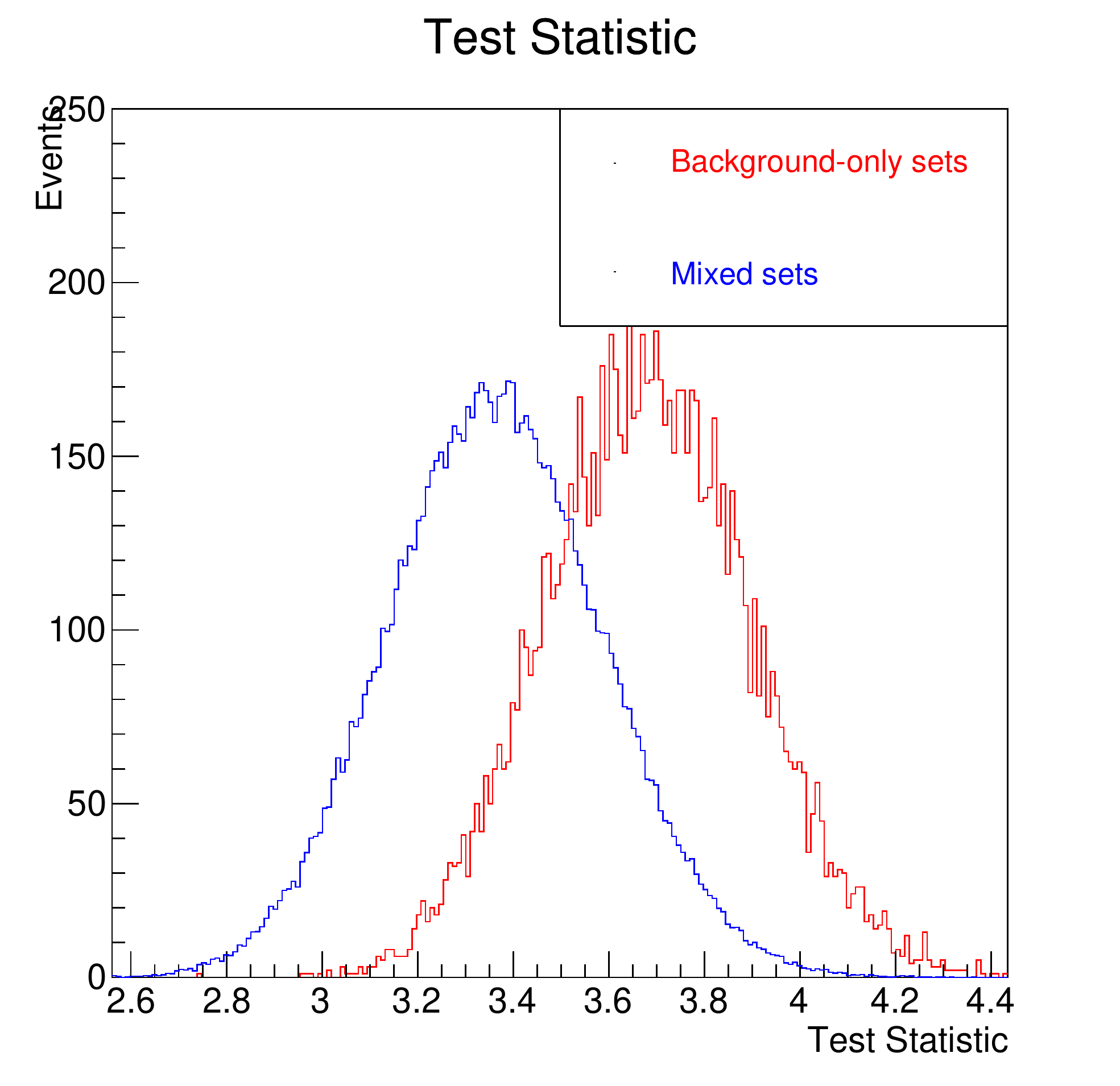}\\
  \caption{{\it Left:} ROC curve expressed in terms of the purity of the discriminators as a function of their efficiency.
    {\it Right:} The values of the test statistic for the background-only sets and for the mixed signal+background sets.}
  \label{fig:1}       
\end{figure}

As a baseline, the pure background training set is populated by 5000 events, whereas the mixed set is composed by 1000 events, with a background fraction $\mathcal{B} = 0.96$. The bootstrap procedure is performed in order to extract 100k subsets of $M=100$ events each. The multi-dimensional goodness-of-fit test described in Sec.~\ref{sec:inverse_bagging} is chosen as a test statistic, and the events are ranked by the average valuee of the test statistic on each bootstrap sample. A direct comparison with event-based classifiers is performed, by using the relative likelihood and the k-Nearest-Neighbour methods discussed in Sec.~\ref{sec:inverse_bagging}. Figure~\ref{fig:1} ({\it left}) shows the Receiver Operating Characteristic (ROC) curve obtained by checking the purity of the discriminator as a function of its efficiency: the Inverse Bagging classifier outperforms both reference classifiers, particularly for high efficiencies. Figure~\ref{fig:1} ({\it right}) shows the values of the test statistic for the background-only sets and for the mixed signal+background sets; the distributions exhibit a good level of separation, indicating that the Inverse Bagging procedure is indeed able to discriminate between background-only sets and mixed signal+background sets.

\begin{figure}[h]
    \centering
    \includegraphics[width=0.45\linewidth,clip]{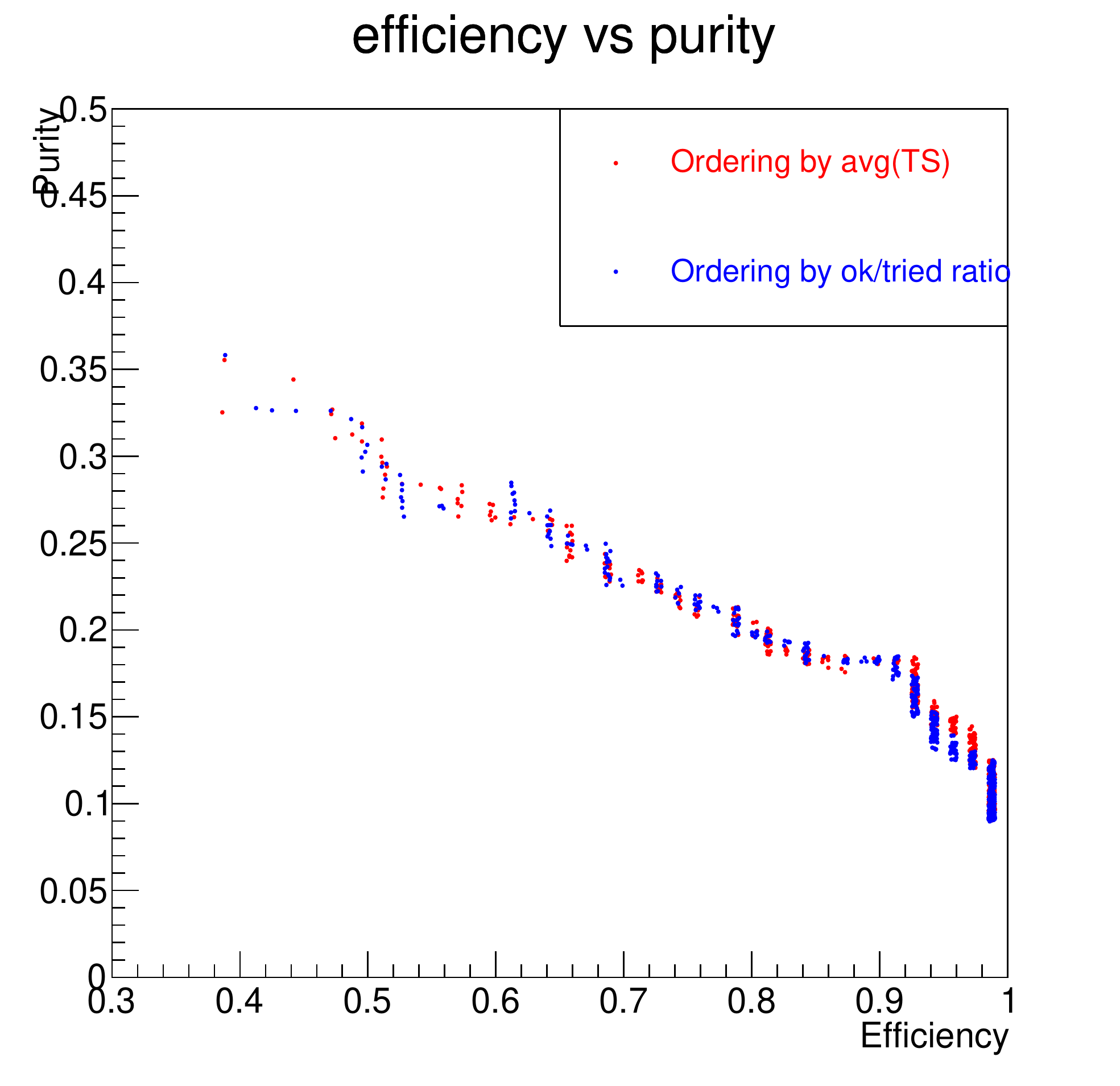}
    \includegraphics[width=0.45\linewidth,clip]{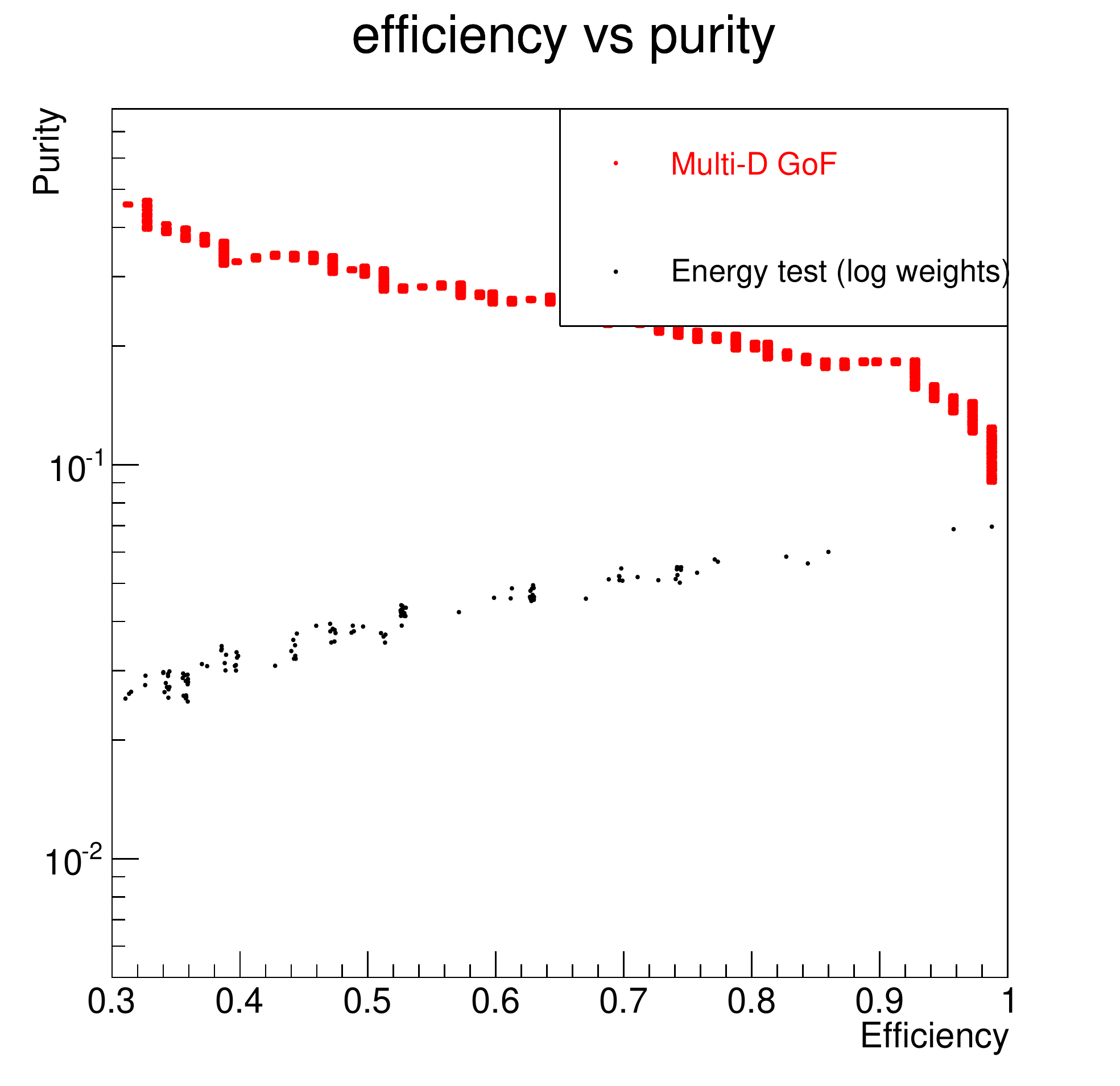}\\
    \caption{{\it Left:} ROC curve for different ordering principles for the progressive removal of bootstrapped sets.
      {\it Right:} ROC curve for different choices of test statistic.}
    \label{fig:2}       
\end{figure}

The simple ratio $R_{o/t}$ ratio, defined in Eq.~\ref{eq:oktriedratio}, between the number of times each test event is picked up to be part of a background-like subset and the number of times the event is picked up to be part of a subset, described in the previous section, is taken as an alternative ordering principle. Figure~\ref{fig:2} ({\em left}) shows that the performance of the Inverse Bagging algorithm is not sensitive to reasonable variations in the ordering principle, at least in this case study.

Zech's energy test, with logarithmic weighting function, is used as an alternative test statistic to check the sensitivity of the Inverse Bagging algorithm to a change in test statistic. Figure~\ref{fig:2} ({\em right}) shows that the performance of the Inverse Bagging algorithm depends heavily on the choice of test statistic. Additional tests with different weights for the energy test, as well as with different test statistics, have been performed, yielding for this dataset worse performances than the multi-dimensional g.o.f. test.

\begin{figure}[h]
  \centering
  \centering \includegraphics[width=0.45\linewidth,clip]{EvsP_base.pdf}
  \centering \includegraphics[width=0.45\linewidth,clip]{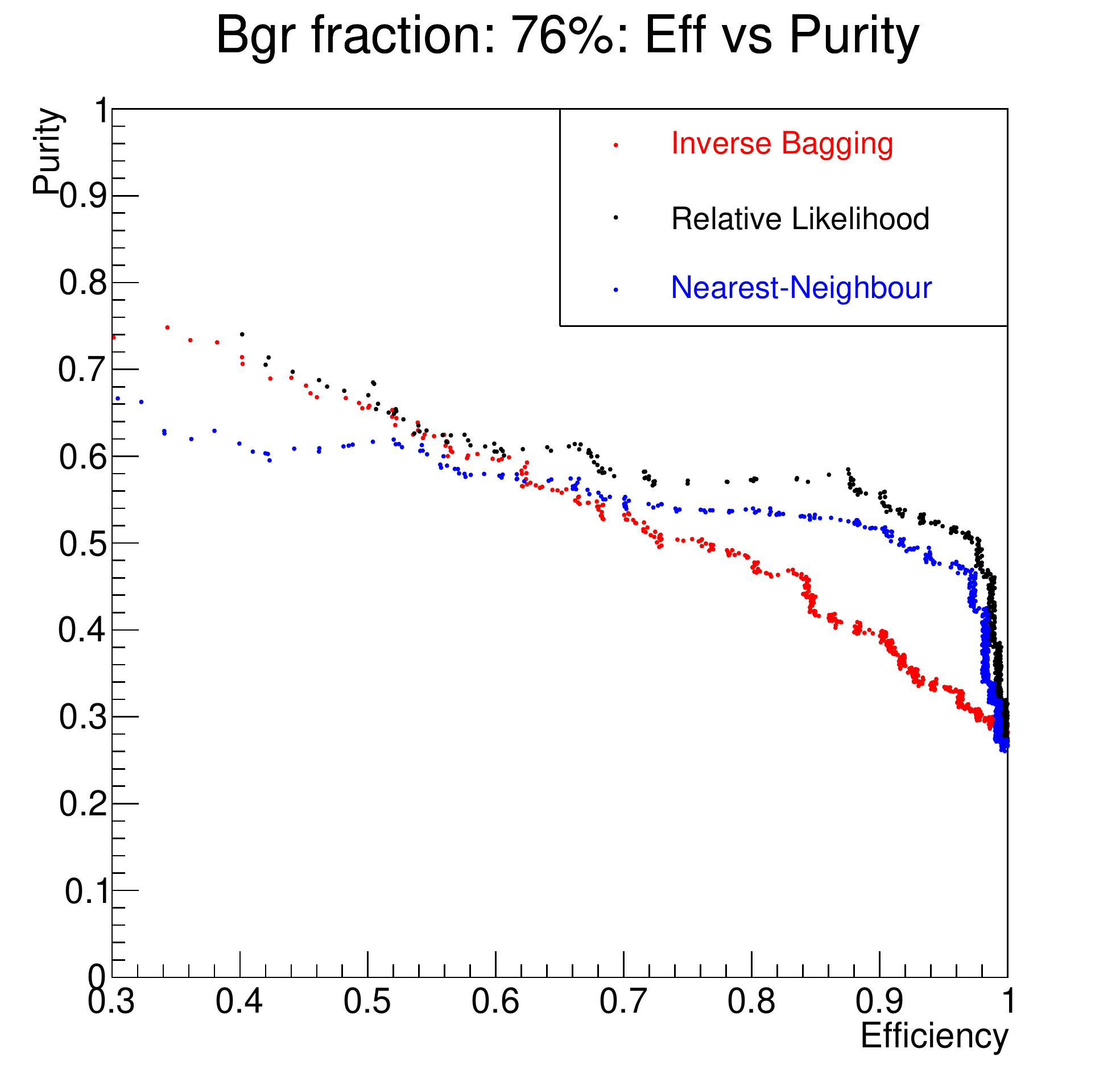}\\
  \caption{{\it Left:} ROC curve obtained for a test sample characterized by a background fraction $\mathcal{B}=0.96$.
    {\it Right:} ROC curve obtained for a test sample charactedized by a background fraction $\mathcal{B}=0.76$}
  \label{fig:3}       
\end{figure}

The Inverse Bagging algorithm is designed to perform well when the composition of the test sample is heavily dominated by one of the two classes: in this case study, this translates to a very high background fractions, $\mathcal{B}=0.96$. An alternative test set, characterized by a lower background fraction, $\mathcal{B}=0.76$, is processed, and the performance of the Inverse Bagging algorithm, as well as the one of the reference classifiers, is evaluated. As expected, the performance of the Inverse Bagging algorithm degrades for less rare signals, up to the point of being worse than the standard reference classifiers; this happens because the main assumption the Inverse Bagging algorithm is based on -- i.e. that it is always possible to pick a significative fraction of bootstap samples composed ony by background events -- no longer holds, whereas the reference classifiers are fed more signal events to pick features from. The result of the check is shown in Fig.~\ref{fig:3}.

\begin{figure}[h]
  \centering
  \includegraphics[width=0.45\linewidth,clip]{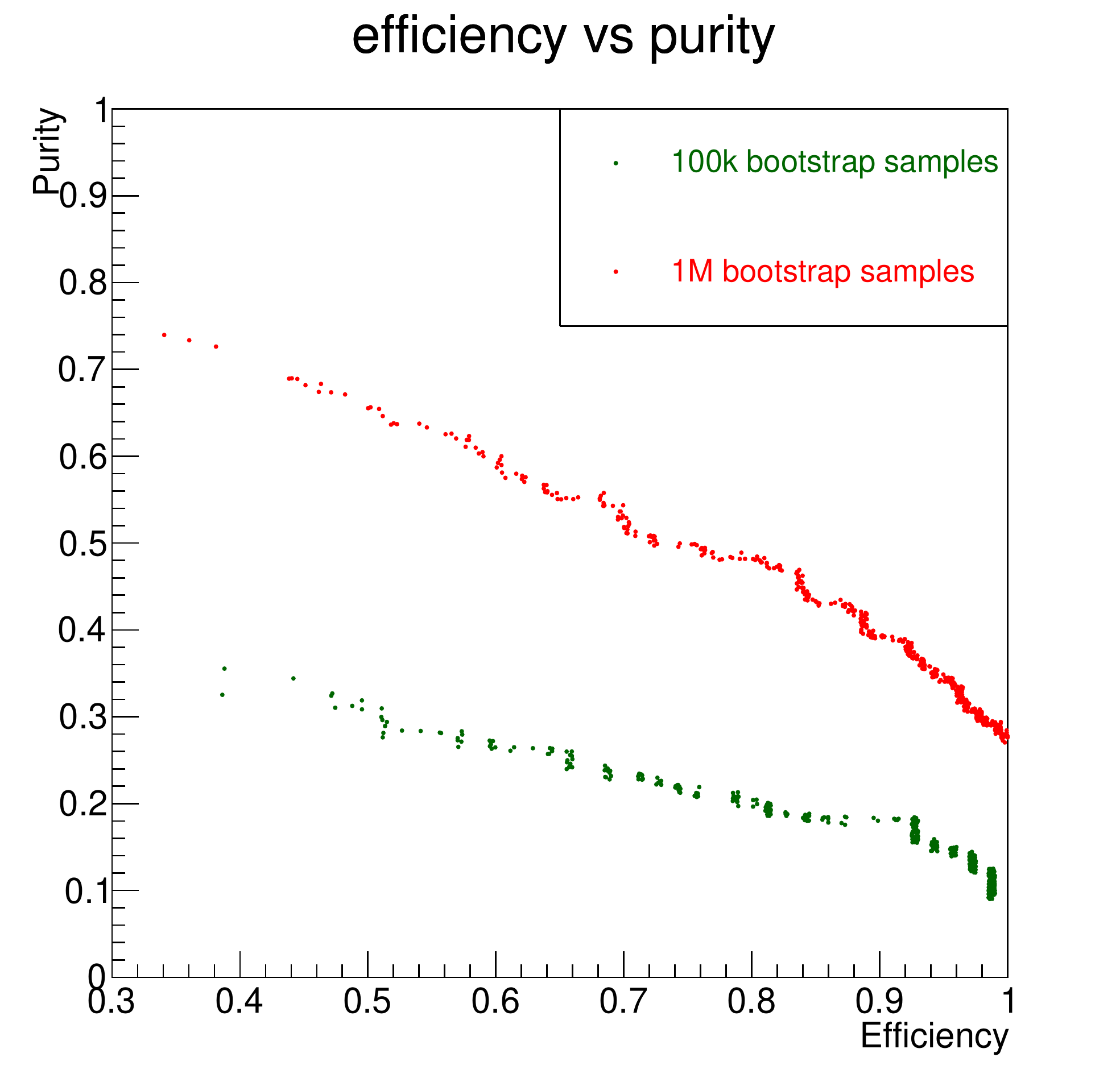}
  \includegraphics[width=0.45\linewidth,clip]{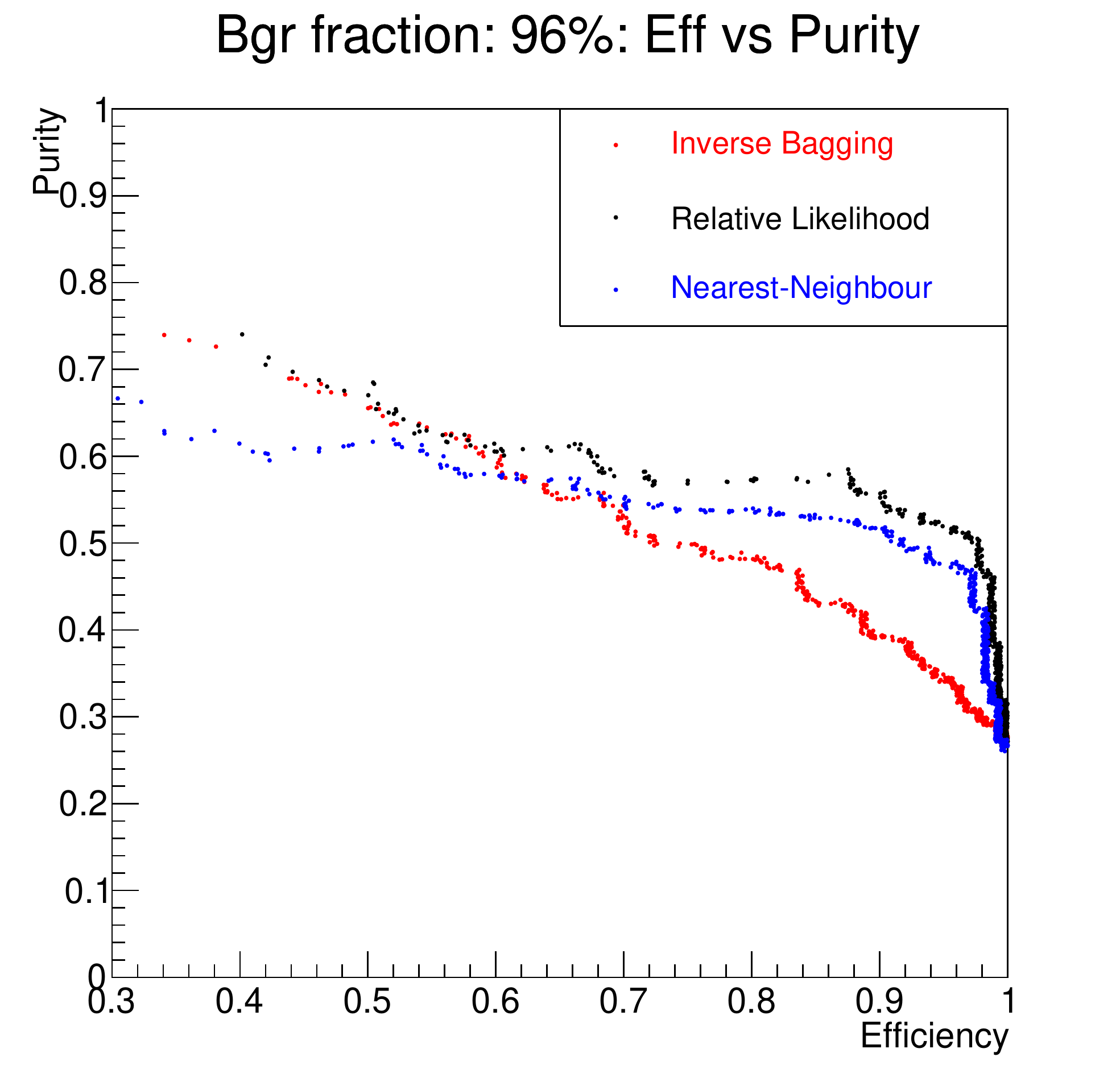}\\
  \caption{{\it Left:} ROC curve for different number of bootstrap samples.
    {\it Right:} Comparison between the Inverse Bagging classifier obtained by using $10^{6}$ bootstrap samples and the reference classifiers.}
  \label{fig:4}       
\end{figure}

Increasing the number of bootstrap samples is shown to have a large impact on the performance of the Inverse Bagging classifier. Figure~\ref{fig:4} ({\em left}) compares, for the baseline background fraction, the performance of the classifier when the number of bootstrap samples is varied from 100k to 1 million; the performance is sensibly higher when the number of bootstrap samples is higher. However, a comparison with the performance of the reference classifiers, shown in Fig.~\ref{fig:4} ({\em right}) suggests that increasing too much the number of bootstrap samples results in a degraded performance with respect to the reference classifiers.

\begin{figure}[h]
  \centering
  \includegraphics[width=0.45\linewidth,clip]{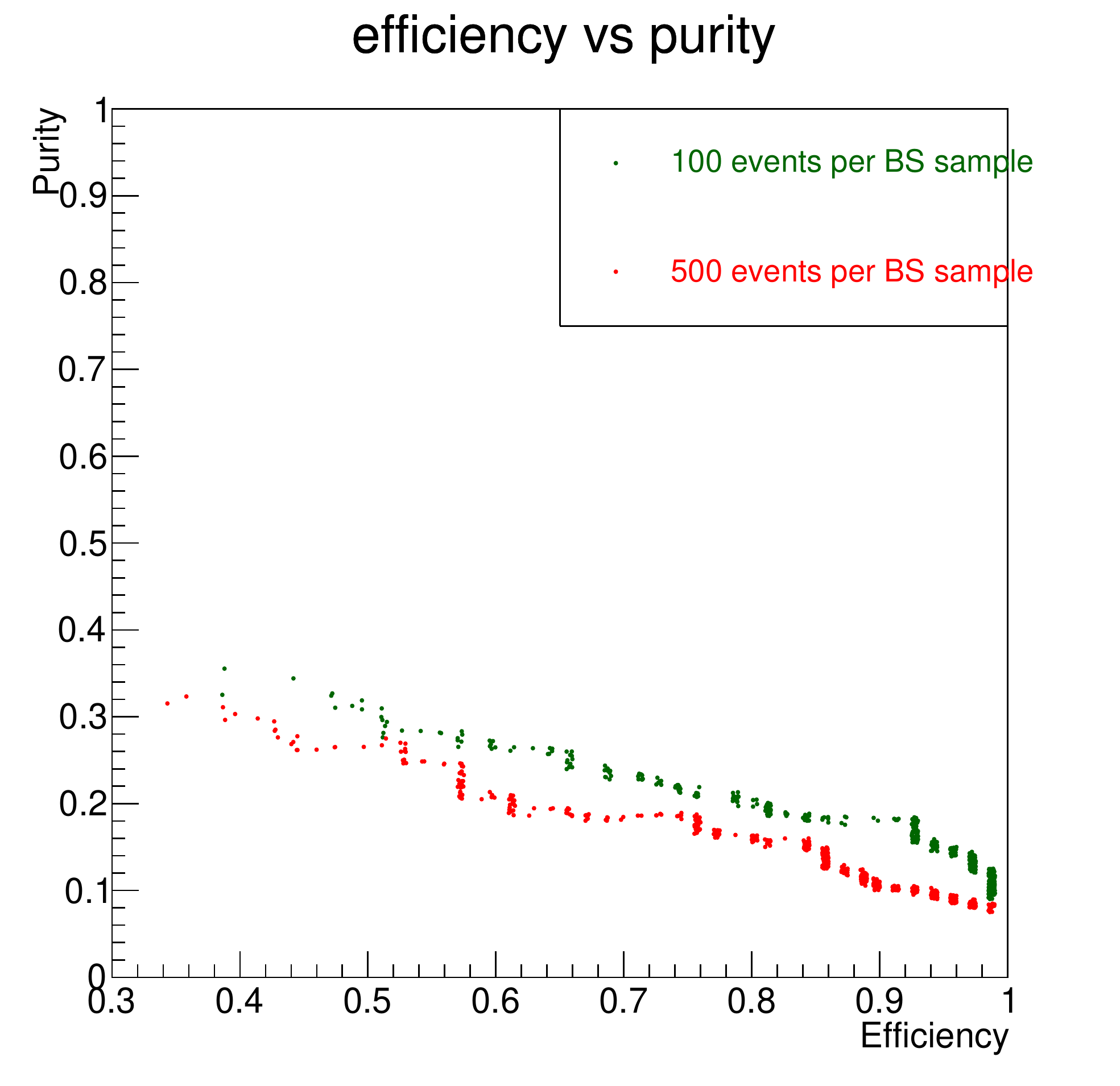}
  \includegraphics[width=0.45\linewidth,clip]{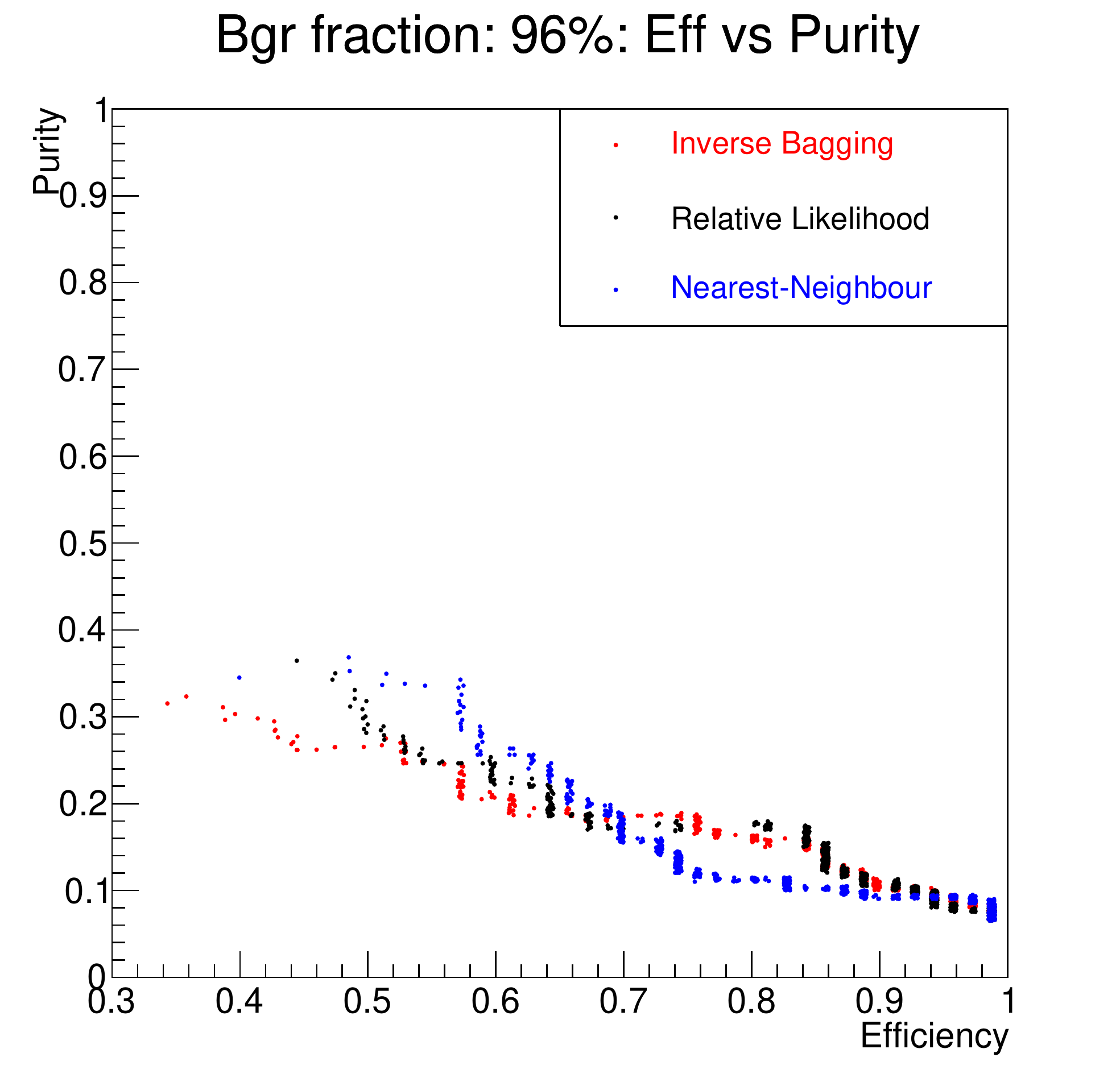}\\
  \caption{{\it Left:} ROC curve for different sizes of the bootstap samples.
    {\it Right:} Comparison between the Inverse Bagging classifier obtained by using $M=500$ events for each bootstrap sample, and the reference classifiers.}
  \label{fig:5}       
\end{figure}

The size $M$ of each bootstrap sample is another parameter of the algorithm that is expected to have an impact on its performance. If $M$ is too large with respect to the size $N_{test}$~of the test sample, the bootstrap procedure would produce too many subsets with a substantial number of events in common, resulting in them having artificially the same value of test statistic and hence the same classification score. Figure~\ref{fig:5} ({\em left}) shows that indeed the performance is degraded when the size of the bootstrap samples is increased from 2\% to 10\% of the size of the test sample. Figure~\ref{fig:5} ({\em right}) suggests that in such a situation the Inverse Bagging algorithm performance might become equal or worse to the one of the benchmark classifiers. However, more detailed studies are needed to extrapolate the general behaviour of the algorithm in terms of number of bootstrap samples.

\section{Summary and future work}
\label{sec:summary_future}

The bootstrap method is a powerful handle for cases in which the underlying PDF is not accessible. Aggregating the information coming from bootstrap results in multivariate classifiers that outperform the basic ones.

A bootstrap technique is designed for scenarios characterized by one large and well modeled background, and one small and unknown signal. Sub-samples, rich in the well modeled process, are gathered via the bootstrap method, and each event is classified according to the frequency for the event to be part of such sub-samples. The resulting algorithm is labeled {\it Inverse Bagging}; its most striking feature is that the event-based classification is inferred from sample-based properties . Comparisons with two benchmark event-based multi-variate classifiers show that the Inverse Bagging algorithm can outperform them in a common scenario from LHC physics. The properties of the Inverse Bagging algorithm have been studied, in terms of dependence of the performance from the core parameters of the algorithm. Additional tests of the asymptotic properties of the algorithm are ongoing, and special care is being dedicated to improving the multi-dimensional test-statistic that already outperforms the other reference test statistics used as reference.

The Inverse Bagging algorithm has been successfully tested on a sample scenario taken from High Energy Physics. Such scenarios are common in LHC new physics searches, and have assumed more and more importance in the recent years, because of the very noisy nature of the data due to the difficult experimental conditions of a high-luminosity hadron collider. The Inverse Bagging algorithm can be used for detecting anomalous or outlier events, which is an important problem for the treatment of datasets dominated by noise events.

%
%
%
%
%

\begin{thebibliography}{}
%
%
\bibitem{EfronTibshirani}
  B. Efron, R.J. Tibshirani, \textit{An Introduction to the Bootstrap} (Springer, Dordrecht, 1993) 31-38, 45-85
\bibitem{Rice}
  J.A. Rice, \textit{Mathematical Statistics and Data Analysis} (Thomson Brooks/Cole, Belmont, 2007) 202-254


\bibitem{Narsky:2005hn}
  I. Narsky, Statistical Problems in Particle Physics, Astrophysics and Cosmology \textbf{PHYSTAT 05 Proceedings}, 143-146 (2005)

\bibitem{Aslan:2002cn}
  B. Aslan, G. Zech, \textbf{arXiv preprint 0203010} (2002)

\bibitem{Baldi2016}
  P. Baldi, K. Cranmer, T. Faucett, P. Sadowski, D. Whiteson, The European Physical Journal C \textbf{76, 5}, 235 (2016)

\bibitem{Altman}
 N.S. Altman, The American Statistician \textbf{46 (3)} 175–185.(1992). 
  

\end{thebibliography}
%

\section{Acknowledgements}

P.V. is funded by the MINECO (Spain) under the project \texttt{MINECO-15-FPA2014-55296-R}.
This work is part of the activities of the Horizon 2020 project ``AMVA4NewPhysics'' (projectID: 675440)

\end{document}